\documentclass{article}

\usepackage{arxiv}

\usepackage[utf8]{inputenc} 
\usepackage{amsmath} 
\usepackage{amsfonts}       

\usepackage{bm}
\usepackage{todonotes}
\usepackage{subcaption}
\usepackage{hyperref}
\usepackage{tabularx}
\usepackage{xcolor}
\usepackage{soul}
\usepackage{graphicx}
\usepackage[square,comma,numbers]{natbib}
\usepackage{bbm}   

\title{
	Using Simulation Optimization to Improve Zero-shot Policy Transfer of Quadrotors\thanks{Published in IEEE Explore: \url{https://ieeexplore.ieee.org/document/9981229} \copyright IEEE 2022.}
}

\author{
	Sven Gronauer\thanks{Contact author: {\tt\small sven.gronauer@tum.de}}, \,Matthias Kissel, Luca Sacchetto, Mathias Korte and Klaus Diepold
	\\ 
	Department of Electrical and Computer Engineering\\ 
	Technical University of Munich (TUM), Germany\\ 
	Arcisstr. 21, 80333 Munich
}

\begin{document}

\maketitle

\begin{abstract}
In this work, we propose a data-driven approach to optimize the parameters 
of a simulation such that control policies can be directly transferred from 
simulation to a real-world quadrotor. Our neural network-based policies take 
only onboard sensor data as input and run entirely on the embedded hardware. 
In extensive real-world experiments, we compare
low-level Pulse-Width Modulated control with higher-level control 
structures such as Attitude Rate and Attitude, which utilize 
Proportional-Integral-Derivative controllers to output motor commands. 
Our experiments show that low-level controllers trained with reinforcement 
learning require a more accurate simulation than higher-level control policies.
\end{abstract}

\section{INTRODUCTION} \label{sec:introduction}


Programming intelligent control strategies for complex robot systems is a 
challenging task. 
Reinforcement learning (RL) promises the automated generation of control 
strategies through a data-driven approach instead of explicitly designing 
hand-crafted solutions through expert knowledge. 
In recent years, the field of RL has witnessed outstanding successes and 
raised a surge of interest in the control of dynamical systems through such a 
trial-and-error paradigm. 
The combination of RL and deep learning methods excel at problems that can 
be quickly simulated like robotics \cite{Lillicrap2016ContinuousControl} 
and video games \cite{mnih2015human} or in domains where the exact model is 
known but long-horizon planning is not computationally tractable, 
e.g. board games like Go and Chess \cite{Silver2018AlphaZero}. 

Despite the significant advances in recent years, the applicability of RL 
algorithms is still limited when the data at test time differ from those seen 
during training \cite{kirk2021survey}. 
Since many real-world systems cannot afford to learn policies from scratch due to the expense of data, simulations are the preferred approach to build data-driven control policies in the RL community. 
Naturally, a gap between the simulation and the real world still exists because 
modeling of all effects either requires in-depth expert knowledge or is 
simply not desirable, e.g. calculating all aero-dynamical effects can 
significantly increase simulation time. A successful policy transfer thus 
requires the reality gap to be small.

In this paper, we address the sim-to-real gap in the domain of quadrotor 
control and investigate three different structures for quadrotor control. 
Our contributions are:
\begin{enumerate}
	\item We apply simulation optimization as a data-driven approach to narrow 
the reality gap and demonstrate that zero-shot policy transfers become feasible 
with minimal tuning effort. 
The data used for optimization were collected for 
approximately one hour on the real-world quadrotor with a pre-implemented 
Proportional-Integral-Derivative (PID) controller of the drone platform.

	\item We deploy low-level control policies based on Pulse-Width Modulated (PWM) 
thrust commands trained entirely in simulation on the quadrotor 
without fine-tuning the policy. To the best of our knowledge, 
this is the first work using RL that accomplishes successful zero-shot 
transfers to a real-world quadrotor based on low-level control while having 
access to only onboard sensor data and running all computations on the 
embedded hardware.
			
	\item We study three control structures that differ in their level of 
abstraction: PWM, Attitude Rate and Attitude control. Through extensive 
real-world experiments, we investigate the required fidelity of the 
simulation with respect to the deployed control structure. 
\end{enumerate}

Throughout this work, we focus on the context of zero-shot policy transfer, 
i.e. we train the agent entirely in simulation and then deploy the controller on 
a quadrotor robot without using real-world data to fine-tune the policy.

\section{RELATED WORK} \label{sec:related-work}

\subsection{Quadrotor Control} 

For attitude control and set-point tracking of quadrotors, a common approach 
is a hierarchical control that consists of nested PID controllers 
\cite{Mahony2012QuadrotorTutorial}. 
Approaches like the linear quadratic regulator are also suitable methods 
for the stabilization around the hover conditions under reasonably small roll 
and pitch angles. However, when more dynamic flight behaviors are desired, 
more complex controllers may be required \cite{Mellinger2011MinimumSnap}.
Due to their highly dynamic movement capabilities, quadrotors depict an 
interesting platform to test maneuvers such as 
landing \cite{Kooi2021InclinedLanding} 
and perform acrobatic maneuvers like loopings and rolls \cite{kaufmann2020RSS}, 
multi-flips \cite{Lupashin2009SimpleLearningStrategy} 
or flying through narrow vertical gaps \cite{Mellinger2011MinimumSnap}.
While most of the aforementioned works rely on highly accurate state estimation, 
only a few papers have considered quadrotor control based on lone onboard sensor 
signals. In \cite{Lambert2019LowLevelControl}, model-based RL was used to train 
a policy from real-world data while relying solely on onboard sensor 
measurements to run control with direct motor PWM signals. 
However, neural network calculations were executed on a server and then 
transmitted via radio to the drone.	They leveraged $180$s of real-world 
flights to produce a controller capable of accomplishing at most $6$s of 
flight time. 

\subsection{Bridging the Reality Gap} 

To reduce the gap between simulation and real-world, 
\textit{domain randomization} has been proposed to augment the diversity of 
data by randomization. One way is to randomize the dynamics where simulation 
parameters responsible for the realization of the system are 
re-sampled at the beginning of every trajectory \cite{Peng2018Sim2Real}. 
Another approach is to randomize the rendering of image-based observations in 
the simulation \cite{Tobin2017DomainRandomization}. 
In \cite{loquercio2019droneracing} it was shown that agile drone racing is 
possible with convolutional neural networks when trained on an 
abundance of image-based data. 
Besides domain randomization, another method to narrow the reality gap is 
\textit{simulation optimization} \cite{Carson1997SimOpt} 
which is a data-driven approach to identify simulation parameters based on 
real-world data. In \cite{Chebotar2019ClosingSim2Real} a simulation parameter 
distribution was learned based on collected data from a physical manipulator robot.

\subsection{Zero-shot Policy Transfer}

In the area of RL, there have been only a few works that study the zero-shot 
policy transfer to real-world quadrotors. \cite{kaufmann2020RSS} showed that 
Attitude Rate control based on image-based and onboard data could perform highly 
agile maneuvers such as rolls and loopings. Most similar to our paper is 
\cite{Molchanov2019Sim-to-Multi-Real} were it was shown that low-level PWM-based 
controllers could stabilize and generalize to multiple sizes of quadrotors 
while entirely being trained in simulation. In contrast to our work, they used 
an external tracking system to accurately estimate the drone state.

\section{PRELIMINARIES} \label{sec:preliminaries}

\subsection{Reinforcement Learning} 	\label{sec:reinforcement-learning}

A Markov Decision Process (MDP)  is formalized by a tuple 
$\left( \mathbb{S}, \mathbb{A}, \mathcal{P}, r, \mu \right)$, 
where $\mathbb{S}$ and $\mathbb{A}$ denote the state and action spaces. 
$\mathcal{P}: \mathbb{S} \times \mathbb{A} \rightarrow P(\mathbb{S})$ 
describes the system transition probability, $\mu$ denotes the initial state 
distribution and $r: \mathbb{S} \times \mathbb{A} \rightarrow \mathbb{R}$ is 
the reward function. 
Let $\tau = (\bm s_0, \bm a_0, \bm s_1, \bm a_1,\dots)$ be a trajectory 
generated under the policy $\pi$ with 
$\bm s_{t+1} \sim \mathcal{P}(\cdot | \bm s_t, \bm a_t), \bm a_t \sim \pi(\cdot | \bm s_t)\, \text{and}\,  \bm s_0 \sim \mu$. 
We apply the shortcut $\tau \sim \pi$  when trajectories are generated under 
$\pi$ and denote the trajectory return by 
$R(\tau) = \sum_{t=0}^\infty \gamma^t r(\bm s_t, \bm a_t)$ 
with the discount factor $\gamma \in (0, 1)$.
In RL, the goal of the agent is to learn 
a control policy $\pi : \mathbb{S} \rightarrow \mathcal{P}(\mathbb{A})$ that 
maximizes the expected return $ J(\pi) = \mathbb{E}_{\tau \sim \pi} [ R(\tau)]$.
We use $\pi_\theta$ to denote that the policy is parametrized by a neural 
network with weights $\bm\theta$.

In this work, we sample system parameters $\bm\xi$ from a distribution $\Xi$ at the beginning of each trajectory. This results in the system transition probability being dependent on the system parameters $s_{t+1} \sim \mathcal{P}(\cdot | s_t, a_t, \xi)$ which is known as domain randomization.
Thus, we seek policy parameters that maximize the expected return 
\begin{align}
\max_{\bm\theta}  J(\pi_\theta) =  \mathbb{E}_{\bm\xi \sim \Xi} \left[ \mathbb{E}_{\tau \sim \pi_\theta} [ R(\tau)] \right]
\end{align}
over the dynamics induced by the distribution of simulation parameters.

\subsection{Quadrotor}		\label{sec:quadrotor}

\begin{figure}[!t]	
	\center
	\includegraphics[width=0.6\textwidth]{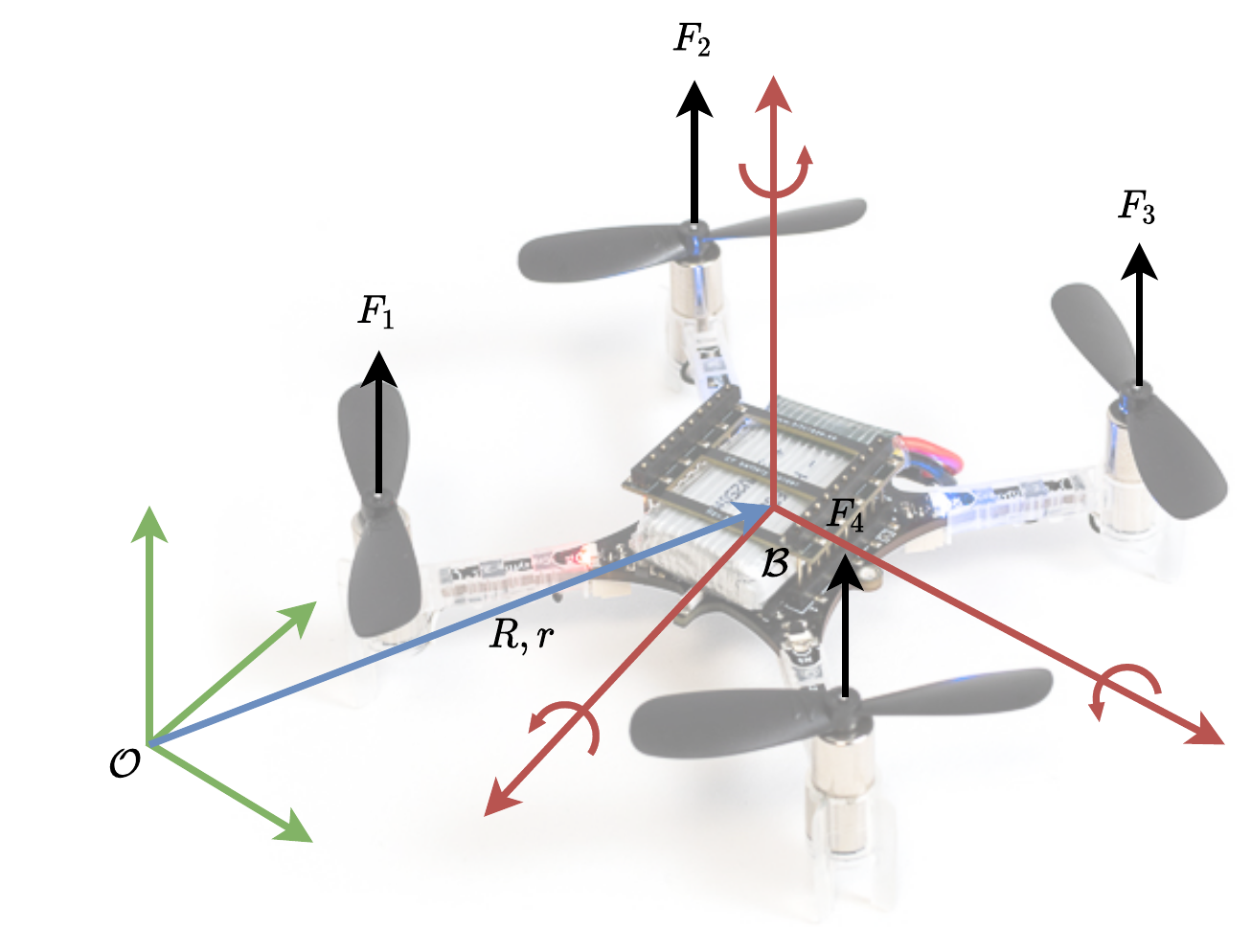}
	\caption{Overview and coordinates frames of the \textit{CrazyFlie} quadrotor.}
	\label{fig:crazy-flie}
\end{figure}		

\subsubsection{Dynamics Model} \label{sec:dynamics-model}

The dynamics of the quadrotor are modeled by the differential equation
	\begin{align}   \label{eq:discrete-system-diff-eq}
    \bm{\dot{x}} = f(\bm{x}, \bm{u})
	\end{align}
where the drone state $\bm{x} = [\bm{r}^T, \bm{\dot r}^T, \bm{\varphi}^T, \bm{\omega}^T]^T \in \mathbb{R}^{13}$ encompasses the position $\bm{r}$, the linear velocity $\bm{\dot r}$, the body angle $\bm{\varphi}$ in quaternions and the angular speed $\bm{\omega}$. 
The acceleration of the drone's center of mass is described by Newton's equation
	\begin{align}   \label{eq:linear-acceleration}
    m\ddot{\bm{r}} = \begin{bmatrix} 0 \\ 0 \\ -mg \end{bmatrix} + R \begin{bmatrix} 0 \\ 0 \\ \sum F_i \end{bmatrix}
	\end{align}
in the inertial frame $\mathcal{O}$ with gravity $g$. The quadrotor mass is 
$m$ and $\sum F_i$ is the sum of the vertical forces acting on the rotors. 
$R$ is the rotation matrix from the body frame $\mathcal{B}$ to the inertial 
frame. The angular acceleration governed by Euler's rotation equations 
in $\mathcal{B}$ is
	\begin{align}   \label{eq:angular-acceleration}
    I \bm{\dot{\omega}} = \bm \eta - \bm{\omega} \times (I \bm{\omega})
	\end{align}
with the inertia matrix $I$. The torques $\bm \eta$ acting in the body frame 
are determined by 
	\begin{align}   \label{eq:torque}
	\bm{\eta} =\begin{bmatrix} 
		\frac{1}{\sqrt{2}}L(-F_1-F_2+F_3+F_4) \\ 
		\frac{1}{\sqrt{2}}L(-F_1+F_2+F_3-F_4) \\ 
		-M_1 + M_2 - M_3 + M_4 
	\end{bmatrix}
	\end{align}
with the rotor forces $F_i$, the corresponding motor torques $M_i$ and the 
drone's arm length $L$.
An overview of the quadrotor setup and the coordinate frames is depicted 
in Figure~\ref{fig:crazy-flie}.

\subsubsection{Motor Model}	\label{sec:motor-model}

The angular speed $\nu_i$ of rotor $i$ produces a vertical force 
	\begin{align}   \label{eq:motor-force}
    F_i =\frac{mg}{4} k_F \nu_i^2
	\end{align}
that lifts the quadrotor. We use $k_F \in \mathbb{R}$ to denote the 
thrust-to-weight ratio. The rotors also produce a moment according to
	\begin{align}   \label{eq:motor-torque}
    M_i = k_{M_1} F_i + k_{M_2}
	\end{align}
with $k_{M_1}$ and $k_{M_2}$ being scalars.
The motor speeds are normalized $\nu_i \in [0, 1]$ and are modeled 
$\bm{\nu} = \sqrt{\bm{u}}$ as the 
square root\footnote{The CrazyFlie firmware internally uses a quadratic model that considers battery level and the 	commanded thrust $\bm{u} \in [0, 1]$ and maps it to the corresponding PWM value.} 
of normalized commanded thrusts $u_i \in [0, 1]$. 
For PWM control, the relationship between the action $\bm a$ taken by the 
agent and the normalized commanded thrust is 
$u_i = \frac{1}{2} \left[ \min(\max(a_i, -1), 1) + 1\right]$.
We model the motor dynamics with a differential equation of first order 
	\begin{align}   \label{eq:motor-dynamics}
    T_m \dot{\nu_i} = - \nu_i + \sqrt{u_i},
	\end{align}
where $T_m \in \mathbb{R}$ is the motor time constant.
A common approach in related work is to neglect the motor dynamics and assume an 
instantaneous thrust acting on the rotors. However, as we observed in our 
experiments, an accurate actuator model is crucial for a successful sim-to-real 
transfer. Additionally, we model the overall latency $\Delta$ of the system 
which should capture all delays emerging in the hardware. Typically, latency 
arises in the state estimator, in the actuation of the motors and by propagating 
data through the neural network policy.

\section{METHODS} \label{sec:methods}

We aim to find a policy that reliably transfers from the simulation to the 
quadrotor robot without using real-world data to fine-tune the policy. To render 
a zero-shot transfer feasible, we optimize the simulation parameters 
based on collected real-world measurements.

Besides simulation optimization, we want to find an answer to our hypothesis that 
\emph{low-level control structures demand a higher simulation fidelity.}
The intuition behind this assumption is that low-level control requires the policy to 
capture the mapping from drone state to motor commands and thus to understand
to underlying quadrotor dynamics.
High-level control, on the other hand, encourages the agent to learn an abstract 
understanding of the task since only the mapping from the drone state to 
an intermediate space is required. 
The mapping from the policy output space to the motor commands is 
subsequently handled by PID controllers.

The remainder of this section describes the details of our simulation 
environment including the observation and action space, the procedure of the 
simulation optimization using Bayesian optimization and an introduction to the 
three control structures that were deployed and evaluated on the real-world 
quadrotor.

\begin{figure*}[t]
	\center
	\includegraphics[width=0.8\textwidth]{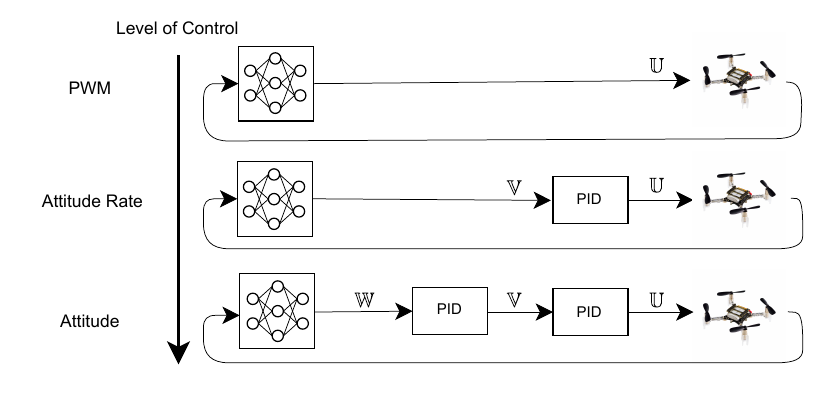}
	\caption{Applied control structures from low to high level.
	(1) \emph{PWM}: A neural network-based policy directly outputs the thrust 
	commands.
	(2) \emph{Attitude Rate}: The policy commands the desired angle rates and 
	collective thrust which are subsequently fed into a PID controller.
	(3) \emph{Attitude}: The cascaded PID controllers take the desired body 
	angle and collective thrust as input and output the commanded thrust 
	for each motor.
	}
	\label{fig:control-structures}
\end{figure*}

\subsection{Simulation Environment}	\label{sec:methods:simulation-environment}

We use the physics simulator PyBullet \cite{coumans2019PyBullet} to calculate 
the dynamics of the quadrotor akin to the work of 
\cite{panerati2021GymPyBulletDrones}. We run the simulator and the motor 
dynamics with $200$Hz while the agent receives noisy observations with 
$\leq 100$Hz depending on the selected control structure. 

\paragraph{Observations} 
The agent perceives the observation 
$\bm s_t = [\bm{x}^T_t, \bm{e}^T_t,\bm a^T_{t-1}]^T \in \mathbb{R}^{20}$ 
containing the drone state vector $\bm{x} \in \mathbb{R}^{13}$, the difference 
vector $\bm{e} = \bm{r} - \bm{t} \in \mathbb{R}^{3}$ between the drone's current 
position and the set-point position and the action 
$\bm a^T_{t-1} \in \mathbb{R}^{4}$ taken previously. Note that the agent has not 
access to all information of the drone state because the rotor speeds cannot 
be measured. To cope with partial observability, the agent is equipped with a 
history of observations with size $H\geq 1$, 
i.e. $(\bm s_{t-H+1},  \dots, \bm s_t) \in \mathbb{R}^{20H}$. 
We conducted a study about the optimal history size $H \in [1,2,4,6,8]$ in 
simulation and found that $H=2$ performed best.

\paragraph{Actions} 
Actions represent abstract, possibly high-level commands which are transformed 
by the control structure to the corresponding motor command 
$\bm u \in \mathbb{U}$. The implementation from $\mathbb{A}$ to $\mathbb{U}$ 
differs for each control structure and is explained in 
Section~\ref{sec:methods:control-strucutures}.

\paragraph{Domain Randomization} 
Throughout training, we randomize the following system parameters uniformly in 
the range $\pm 10\%$ of the default value: Thrust-to-weight ratio $k_F$, physics 
time-step, quadrotor mass $m$, diagonal of inertia matrix $I$, motor time 
constant $T_m$ and yaw-torque factors $k_{m_1}$ and $k_{m_2}$. 

\paragraph{Noise} As sensor noise, we add Gaussian and uniform noise for 
positions $\bm{r}$, velocities $\bm{\dot{r}}$ and angles $\bm{\varphi}$. For the 
angle rates, we apply the sensor model proposed in \cite{Furrer2016RotorS} which 
adds a Gaussian noise and a time-varying bias to $\bm{\omega}$.
In a similar vain to \cite{Molchanov2019Sim-to-Multi-Real}, we use a discretized 
Ornstein-Uhlenbeck process to model actuator noise which we add to $\bm u$.

\subsection{Simulation Optimization} \label{sec:methods:simulation-optimization}

Simulation optimization is an approach to tune the parameters of a simulation 
through data \cite{Carson1997SimOpt}. The aim is to bring the simulation as 
close  as possible to the real-world and ideally close the reality gap. 
The objective function which we want to minimize is
	\begin{align} \label{eq:sim-opt-objective}
	\min_{\bm\xi \in \Xi} F(\bm\xi, \mathcal{D}) &= \sum_t^T \delta^t  \| W \left( \bm x^{sim}_t(\bm\xi) - \bm x^{real}_t \right)\|_1 \nonumber \\&+ \sum_t^T \delta^t  \| W \left( \bm x^{sim}_t(\bm\xi) - \bm x^{real}_t \right)\|_2
	\end{align}
where $T$ is the mini-trajectory length, $\mathcal{D}$ is the data set and 
$\delta \in (0, 1)$ is a factor to discount later stages of the trajectory where 
the divergence increases due to discrepancies between simulation and real-world. 
We weight the drone state difference $\bm x^{sim}_t(\bm\xi) - \bm x^{real}_t$ 
according to the matrix $W$ with the purpose to scale angle errors and angle 
rate errors differently. Similar to \cite{Chebotar2019ClosingSim2Real}, we use 
the sum of L1 and L2 norm. 

Prior sim-to-real work reported that the accurate modeling of the actuators 
including dynamics, noise and delays is crucial for a successful transfer. 
In \cite{Peng2018Sim2Real} it was pointed out that randomizing the latency of 
the controller and injecting sensor noise to the simulation are key components 
for a high success rate. Similarly, it was demonstrated in 
\cite{Hwangbo2019LearningAA} that learned actuator dynamics significantly 
helped to reduce the reality gap. 
Guided by these results, we decided to optimize those simulation parameters 
$\bm\xi = [k_F, T_m, \Delta]^T$ that describe the motor behavior as well as the 
system latency.

Bayesian Optimization (BO) is our method of choice, as it is regarded as a 
good option for the optimization in continuous domains with less than 20 
dimensions and is robust against stochastic function evaluations 
\cite{frazier2018tutorial}.
BO is particularly suited for objective functions that have long evaluation 
times and can cope with low sample complexity. In contrast to gradient-based 
methods, BO is not prone to converge to local optima.

\subsection{Control Structures} 	\label{sec:methods:control-strucutures}

In this work, we investigate the following three control structures, ordered 
from low to high-level:
\begin{enumerate}
	\item \emph{PWM Control:} 
	The agent is supposed to learn a mapping $\pi: \mathbb{S} \rightarrow 
	\mathbb{U}$ from observation space to thrust commands. The agent receives 
	noisy observations from the Kalman state estimator with $100$Hz. 

	\item \emph{Attitude Rate Control:} 
	The policy outputs lie in the space 
	$[c, \bm \omega_d^T]^T \in \mathbb{V} \subseteq \mathbb{R}^4$ 
	which consists of the mass-normalized collective thrust $c$ and the 
	desired body angle rate $\bm \omega_d$. The PID controller calculates 
	the thrust commands $\bm u \in \mathbb{U}$ based on $c$ and the error 
	between the actual $\bm \omega$ and the desired $\bm \omega_d$.
	The agent receives noisy observations with $50$Hz and is supposed to learn the mapping $\pi: \mathbb{S} \rightarrow \mathbb{V}$.

	\item \emph{Attitude Control:} 
	The agent is incentivized to learn the mapping 
	$\pi: \mathbb{S} \rightarrow \mathbb{W}$ where the policy output space 
	$[c, \bm \varphi_d^T] \in \mathbb{W} \subseteq \mathbb{R}^4$ encompasses 
	the mass-normalized collective thrust $c$ and the desired body angle 
	$\bm \varphi_d$. Subsequently, the mapping 
	$\mathbb{W} \rightarrow \mathbb{V} \rightarrow \mathbb{U}$ is handled by 
	two cascaded PID controllers. Noisy observations are fed into the policy 
	with $25$Hz.
\end{enumerate}
These three control structures are illustrated in Figure~\ref{fig:control-structures}.
We test our hypothesis about the fidelity of the simulator based on the employed 
control method in the next section.


\section{RESULTS} \label{sec:results}

In this section, we describe our experiments and discuss the results. 
First, we elaborate on the setup of the hardware and the learning task. 
Second, we explain our simulation optimization experiments with BO and present 
our found simulation parameters. 
Third and last, we describe our zero-shot transfer experiments that were 
conducted on the real-world quadrotor. After training control policies in 
simulation, we studied the three control structures PWM, Attitude Rate and 
Attitude in order to test our control level hypothesis of 
Section~\ref{sec:methods}.

\begin{table*}[!th]
\caption{Simulation optimization results found by BO. 
The mean and standard deviation was calculated over three trials.
}
\centering
\begin{tabularx}
{\textwidth} { | X | c | c | c |  c | c |} 
\hline
Parameter 										& $T=10$ 						& $T=20$						& $T=30$						& $T=40$ 						& $T=50$\\ 
\hline 
Thrust-weight $k_F$	& $1.746\pm0.0050$	& $1.733\pm0.0093$ 	& $1.725\pm0.0018$	& $1.725\pm0.0081$  	& $1.722\pm0.0027$\\
Time Constant $T_m$ 		& $0.102\pm0.0110$  	& $0.087\pm0.0087$	& $0.088\pm0.0006$	& $0.096\pm0.0035$ 	& 	$0.104\pm0.0012$\\
Latency $\Delta$							& $0.006\pm0.0039$	& $ 0.014\pm0.0038$ 	& $0.018\pm0.0003$ 	& $0.018\pm0.0007$	& $0.018\pm0.0004$\\
\hline \end{tabularx}
\label{table:results-sim-opt}
\end{table*}

\subsection{Experimental Setup} \label{sec:experimental-setup}

As quadrotor robot, we used the \textit{CrazyFlie} 
$2.1$\footnote{\url{https://www.bitcraze.io/products/crazyflie-2-1/}}. 
Due to its small size with a motor-to-motor diameter of $13$cm and light weight 
of $28$g, the \textit{CrazyFlie} drone is very agile and demands high control 
frequencies. Furthermore, due to its low-cost design, the \emph{CrazyFlie} 
exhibits high parameter uncertainties in its building parts which additionally 
requires the controller to be robust over a large system parameter range. 
The drone is equipped with a 3-axis gyroscope, a 3-axis accelerometer and a 
z-axis LIDAR. The sensor data were fed into an extended Kalman filter that runs 
with $100$Hz for drone state estimation. For our experiments, we only used 
onboard sensor measurements and did not use an external tracking system.
For evaluation purposes, we transmitted the flight information via 
\emph{CrazyRadio} to a host computer where we analyzed the logged data. 

As learning task, we want the quadrotor to fly a circle figure with a diameter 
of $0.5$m and with a period of $3$s. Each trajectory starts on a random point of 
the reference circle at the height of $1$m going in clock-wise direction. 
The trajectory length in simulation is $500$, which is equivalent to $5$s of 
real-world time. To incentivize the drone following the set-point trajectory 
$\bm t$, the reward function is designed as
	\begin{align}
	r(\bm s, \bm a, \bm t)
	&= \|\bm e\|_2 + 0.0001\| \bm{a_t} \|_2  + 0.001\| \bm{a_{t-1}} \nonumber\\
	&- \bm{a_t}\|_2 + 0.001 \| \bm \omega \|_2 + r_f
	\end{align}
with $r_f = -100$ being the terminal reward if  $\|\bm e\|_2 > 0.25$ else $r_f = 0$.

\subsection{Simulation Optimization} \label{sec:results:simulation-optimization}

We used the pre-implemented cascaded PID position controller from the 
\emph{CrazyFlie} platform to collect data tuples $( \bm x, \bm u)_t$ 
while the drone was flying the circle figure as described in 
Section~\ref{sec:experimental-setup}. Data were logged with 100Hz and we 
collected approximately one hour of real-world data. 
We think that less data would be also sufficient but we did not consider the 
amount of samples in this work.
For building the data set $\mathcal{D}$, we used every tenth data-point from
the collected real-world data as starting state for a mini-trajectory of 
length $T$. For instance, for $T=50$ we obtained a data set of size 
$\mathcal{D} \in \mathbb{R}^{37673 \times 50 \times 13}$. 

With BO, we opted to find the global optimum over the bounded set of system 
parameters $k_F \in [1.5, 2.5]$, $T_m \in [0.01, 0.50]$ and 
$\Delta \in [0.00, 0.05]$ with respect to $\mathcal{D}$. 
We ran the experiment for 250 evaluations on the objective from 
(\ref{eq:sim-opt-objective}) and averaged the results over three trials. 
We tested different mini-trajectory lengths $T \in \{10,20,30,40,50\}$. 
As hyper-parameters, we used $\delta = 0.95$ as the discount factor and as function approximator we used a Gaussian process with the acquisition function uniformly chosen from lower confidence bound, expected improvement and probability of improvement.

The results obtained by the simulation optimization are shown in 
Table~\ref{table:results-sim-opt}. Contrary to our expectation, 
the mini-trajectory length $T$ had only a minor impact on the obtained results. 
Except for the results for $T=10$, the found parameters $\bm \xi$ lie in a 
similar range. 

\subsection{Zero-Shot Experiments}

We conducted zero-shot experiments on the real quadrotor where we first trained 
policy networks with different simulation parameters and thereafter measured the 
transfer performance.
We decided on such experimental setup for three reasons. 
First, we wanted to evaluate the performance of 
the parameters found by simulation optimization for an induced reality gap. 
Second, we intended to study the robustness of the policies trained in 
different scenarios. 
Third and last, we wanted to test our control level hypothesis as stated in 
Section~\ref{sec:methods}.


We aligned the hyper-parameters to the ones suggested in 
\cite{Gronauer2021SuccessfulIngredients} and \cite{henderson2018matters}.
We applied a distributed learner setup where the policy gradients were computed 
and averaged across $64$ workers with a batch size of $64000$. We trained with 
the Proximal Policy Optimization \cite{schulman2017ppo} algorithm over $500$ 
epochs and applied as discount factor $\gamma=0.99$.
For the neural network architecture, we used multi-layer perceptrons with two hidden 
layers. The critic network used 64 neurons each followed by tanh non-linearities 
whereas the actor had 50 neurons in each hidden layer with ReLU as activations. 
Since the policy network is supposed to run on the drone micro-controller, we 
reduced the number of hidden neurons to achieve an inference time of $<1$ms for 
one forward pass. The weights were set according to Kaiming Uniform and biases 
were initialized as zero vectors. We did not apply parameter sharing between the 
actor and the critic. Both networks were optimized with Adam where the learning 
rate of the value network was $0.001$ and $0.0003$ for the policy. To reduce the 
variance of critic estimates, we applied Generalized Advantage Estimation (GAE) 
\cite{Schulman2016GAE} with the weighting factor $\lambda = 0.95$.
Over the training, we used stochastic policies where the policy output is the 
mean of a multi-variate Gaussian distribution 
$\bm a \sim \mathcal{N}(\pi(\bm s),\epsilon I)$ with $I$ 
being the identity matrix and $\epsilon \in \mathbb{R}$ the exploration noise 
that was linearly annealed from $0.5$ to $0.01$. 


\begin{figure*}[t]
     \centering
     \begin{subfigure}[b]{0.32\textwidth}
         \centering
         \includegraphics[width=\textwidth]{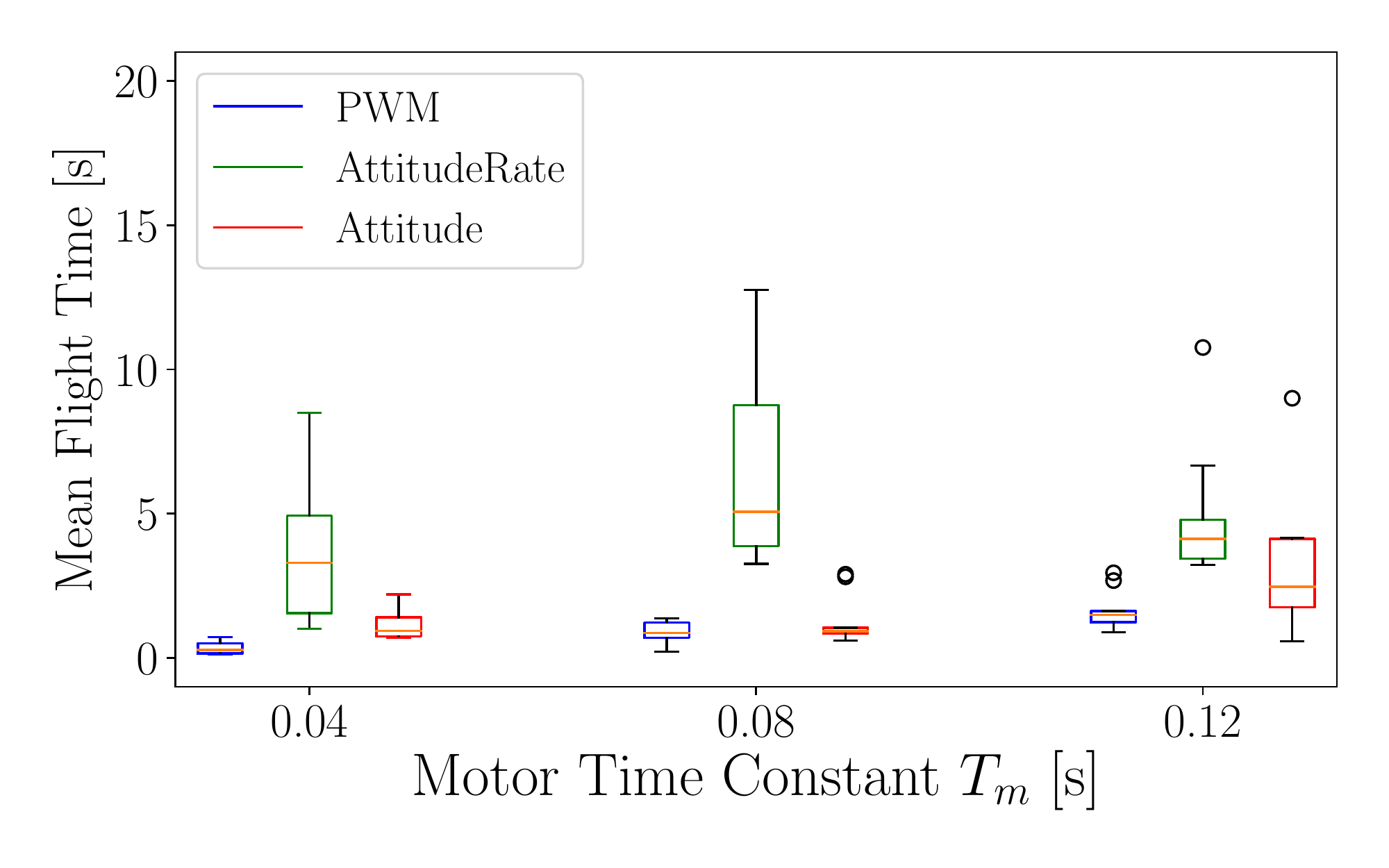}
		\caption{Latency $\Delta=0$ms}
         \label{fig:y equals x}
     \end{subfigure}
     \hfill
     \begin{subfigure}[b]{0.32\textwidth}
         \centering
         \includegraphics[width=\textwidth]{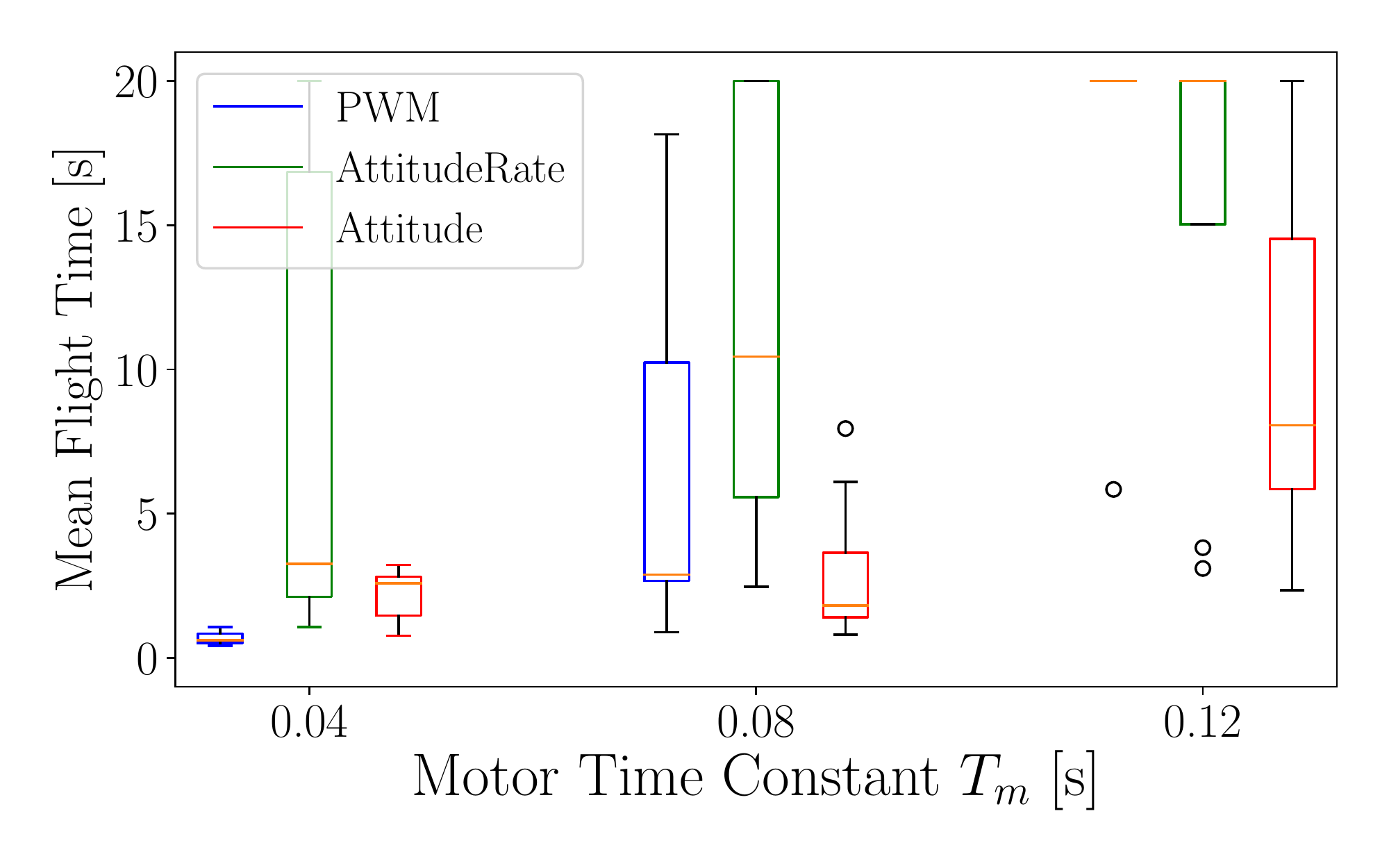}
         \caption{Latency $\Delta=15$ms}
         \label{fig:three sin x}
     \end{subfigure}
     \hfill
     \begin{subfigure}[b]{0.32\textwidth}
         \centering
         \includegraphics[width=\textwidth]{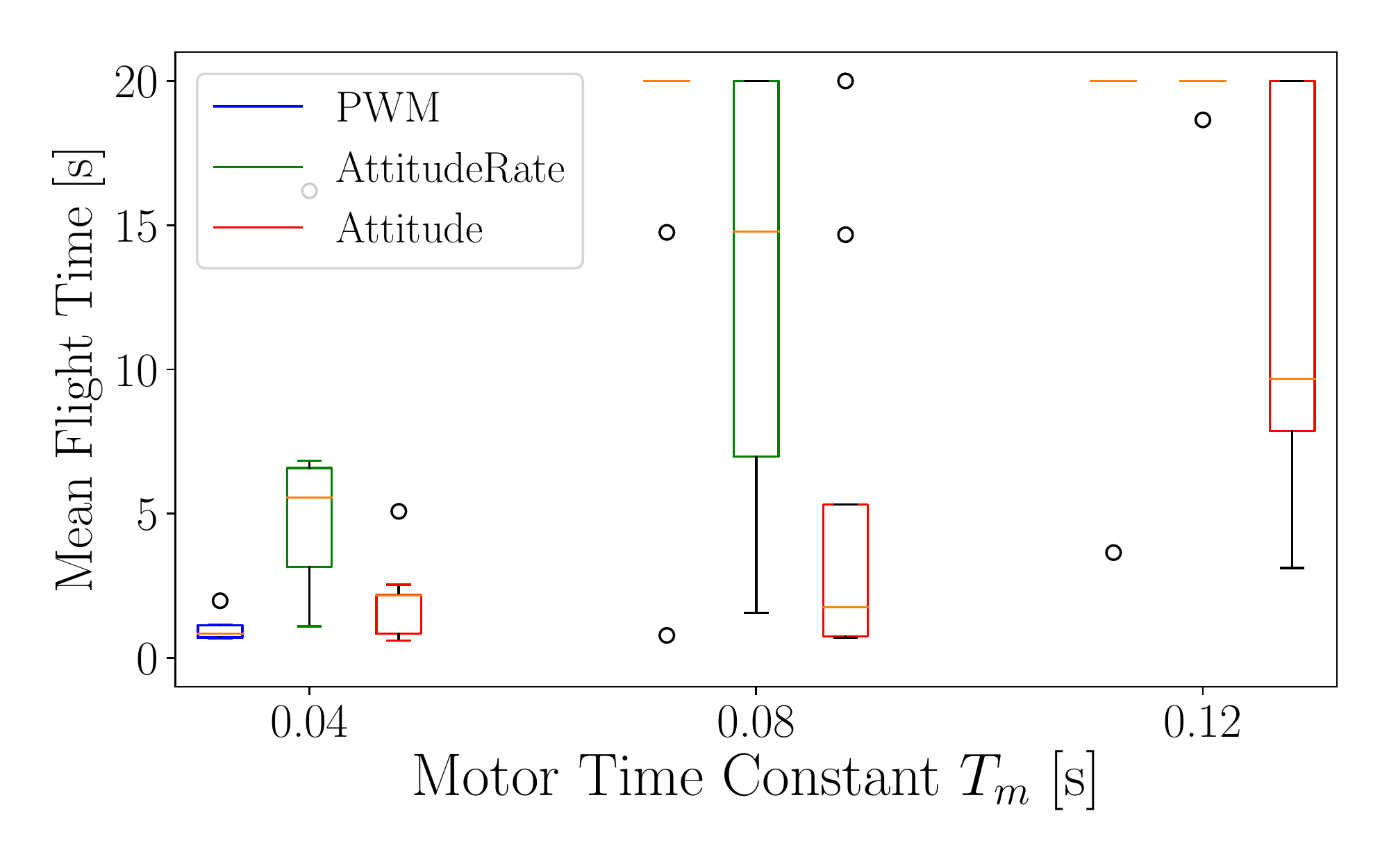}
		\caption{Latency $\Delta=20$ms}
         \label{fig:five over x}
     \end{subfigure}
        \caption{
        Measured flight times in the zero-shot experiments evaluated for different system parameters and control structures. Each box represents nine real-world flights with a maximum flight time of $20$s after which we manually stopped the policy execution. 
        }
        \label{fig:experiments:zero-shot}
\end{figure*}


After training for each combination of motor time 
constant $T_m \in \{0.04, 0.08, 0.12\}$s and latency 
$\Delta \in \{0, 0.015, 0.02\}$s three policies in simulation, we tested the 
policies on the real quadrotor robot.
We applied this evaluation procedure for each control structure and determined 
the flight time of policies over three real-world trials. 
Overall, this resulted in $3^4=81$ trained neural networks and thus $243$ flights. 
If a policy was able to fly longer than $20$s, we manually stopped the 
execution of the neural network. If the policy was not able to stabilize the 
quadrotor, the trial was terminated by a stabilizing backup PID controller, 
which intervened when the roll or pitch angle was greater than $30^\circ$ or when 
roll and pitch rates exceeded $800^\circ$/s. We observed that the rollouts also 
failed when the distance between the drone and the set-point exceeded the 
episode termination criterion of $\|\bm e\|_2 > 0.25$m. The results are 
displayed in Figure~\ref{fig:experiments:zero-shot}.

The PWM control structure showed good zero-shot performance when trained on the 
motor time constant $T_m=120$ms with latency $\Delta \geq 15$ms. The parameter 
setting $T_m=80$ms, $\Delta=20$ms closest to the parameters suggested by the 
simulation optimization resulted in a median flight time of $20$s and two 
outlier flights. However, outside this parameter setting, PWM fastly dropped in 
performance and often failed. When the latency was set to $\Delta=0$ms, PWM yielded the worst performance compared to the other two control approaches.

Overall, the Attitude Rate control structure showed the most robust results over 
the tested system parameters. The best performance was observed for the setting 
$T_m=120$ms and $\Delta=20$ms. In contrast to the other control structures, 
Attitude Rate was also able to stabilize the quadrotor when trained on latency 
$\Delta=0$ms, although most flights terminated early due to the 
$\|\bm e\|_2 > 0.25$m termination criterion.

Surprisingly, Attitude control showed the worst performance despite being the 
highest control level structure. Only the policies that were trained on settings 
with $T_m=120$ms and latency $\Delta \geq 15$ms could achieve a flight duration 
over $10$s.

\section{DISCUSSION} 	\label{sec:discussion}

In this section, we discuss our results from the experiments and draw 
connections to related work. Further, we point out important observations that 
we made during our experiments.

\paragraph{Simulation Optimization}
The system parameters closest to the ones found by the simulation optimization 
showed a reasonable transfer success for the PWM and Attitude Rate control. This 
underlines the potential of data-driven parameter estimation compared to 
classical methods known from system identification. A benefit is the ease of use 
since a handful of system parameters can be estimated simultaneously and no 
isolated measurements of individual physical properties must be conducted.

\paragraph{Control-level Hypothesis}
For low-level PWM control, we found that good zero-shot performance can only be 
achieved when the dynamics of the actuator and the latency are accurately 
estimated. 
However, if the reality gap is too large, low-level policy control is prone to 
fail. The higher-level Attitude rate control, in contrast, shows high robustness 
towards the choice of system parameters. 
This confirms our control level hypothesis that low-level control requires 
higher simulation fidelity than higher-level control structures. Merely Attitude 
control showed disappointing results in the zero-shot experiments, which we 
think is due to the small control frequency of $25$Hz.

\paragraph{Actuator Model}
Related work from quadrupedal robots \cite{Hwangbo2019LearningAA} and robotic 
manipulators \cite{Peng2018Sim2Real} suggested that accurate modeling of the actuators is a crucial component for reliable zero-shot policy transfers. Our experiments confirmed that this statement is also valid for quadrotor robots. In addition to that, we observed that adding latency to the simulation acts as a kind of regularization which helped to improve the zero-shot results.

\paragraph{Robustness}
Due to crashes, we frequently changed spare hardware parts like propellers and motors. 
Although facing a variety of drone parameters, the zero-shot transfers worked 
reliably. Moreover, we tested some of the trained policies also on other
\emph{CrazyFlie} drones which showed similar flight behavior. This indicates 
that the trained policies are robust over a large parameter distribution and 
different drone systems. 

\paragraph{Bang-bang behavior} 
Bang-bang behavior is a known issue in RL where policies prefer an action 
selection towards the boundaries of the action space \cite{Seyde2021bangBang}. 
Without adding penalties to the actions, we observed that agents produced large 
changes in the control outputs which caused a severe performance drop in the 
transfers. By adding penalties for actions $\|\bm a_t \|_2$ and action rates 
$\|\bm a_{t-1} - \bm a_t \|_2$, we were able to conduct reliable transfers to 
the real robot. Similar conclusions were made in sim-to-sim experiments where 
penalties for action and action rate improved both the transfer and the 
robustness towards disturbances \cite{Seyde2021bangBang}. Further, we noticed 
that the transfers failed when the action range was selected too high, e.g. an 
Attitude rate control with the range $[-360, 360]$$^\circ$/s did not work 
whereas the range $[-60, 60]^\circ$/s showed the desired results.


\addtolength{\textheight}{-1.8cm}  

\section{CONCLUSION}	\label{sec:conclusions}

In this paper, we demonstrated that low-level control policies entirely trained 
with reinforcement learning in simulation 
can be successfully transferred to a quadrotor robot when the reality gap is small.
To render such zero-shot policy transfers feasible, we narrowed the sim-to-real gap
by collecting real-world 
data with a pre-implemented PID controller and applying simulation optimization.
Our neural network-based policies used only onboard sensor data for inference 
and ran on the embedded micro-controller of the drone. 
Finally, we conducted extensive real-world experiments and compared three 
different control structures ranging from low-level PWM motor 
commands to high-level Attitude control based on cascaded PID controllers. The
experiments confirm our hypothesis that low-level control policies require a 
higher simulation fidelity.

\section*{APPENDIX}

\begin{table}[!ht]
\caption{Used drone parameters.
}
\centering
\begin{tabularx}
{\columnwidth} { | X | r | r |} 
\hline
Parameter& Value & Physical Unit\\ 
\hline 
Gravitational acceleration $g$											& $9.81$ 				& m$\cdot$s$^{-2}$				\\
Inertial $I_{xx}$								& $1.33 \cdot 10^{-5}$	& kg$\cdot$m$^2$					\\
Inertial $I_{yy}$								& $1.33 \cdot 10^{-5}$	& kg$\cdot$m$^2$						\\
Inertial $I_{zz}$								& $2.64\cdot 10^{-5}$ 	& kg$\cdot$m$^2$						\\
Length $L$										& $0.0396$ 				& m					\\
Mass $m$										& $0.028$ 				& kg					\\
Torque-to-weight ratio $k_{M_1}$	& $5.96\cdot 10^{-3}$ & m	\\
Torque-to-weight ratio $k_{M_2}$	& $1.56\cdot 10^{-5}$	& N$\cdot$m	\\
\hline \end{tabularx}
\label{table:drone-parameters}
\end{table}

\section*{ACKNOWLEDGMENT}

We want to thank Matthias Emde for his support in the implementation of drone 
firmware. 
The project on which this report is based was supported by the German Federal 
Ministry of Education and Research under grant number 01IS17049. The authors are 
responsible for the content of this publication.


\bibliographystyle{plain}
\bibliography{bibliography}

\begin{thebibliography}{10}

\bibitem{Carson1997SimOpt}
Yolanda Carson and Anu Maria.
\newblock Simulation optimization: Methods and applications.
\newblock In {\em Proceedings of the 29th Conference on Winter Simulation}, WSC
  '97, pages 118--126, USA, 1997. IEEE Computer Society.

\bibitem{Chebotar2019ClosingSim2Real}
Yevgen Chebotar, Ankur Handa, Viktor Makoviychuk, Miles Macklin, Jan Issac,
  Nathan Ratliff, and Dieter Fox.
\newblock Closing the sim-to-real loop: Adapting simulation randomization with
  real world experience.
\newblock In {\em 2019 International Conference on Robotics and Automation
  (ICRA)}, pages 8973--8979, 2019.

\bibitem{coumans2019PyBullet}
Erwin Coumans and Yunfei Bai.
\newblock Pybullet, a python module for physics simulation for games, robotics
  and machine learning.
\newblock \url{http://pybullet.org}, 2016--2021.
\newblock Accessed: 2021-12-19.

\bibitem{frazier2018tutorial}
Peter~I. Frazier.
\newblock A tutorial on bayesian optimization.
\newblock {\em CoRR}, abs/1807.02811, 2018.

\bibitem{Furrer2016RotorS}
Fadri Furrer, Michael Burri, Markus Achtelik, and Roland Siegwart.
\newblock {\em RotorS---A Modular Gazebo MAV Simulator Framework}, pages
  595--625.
\newblock Springer International Publishing, Cham, 2016.

\bibitem{Gronauer2021SuccessfulIngredients}
Sven Gronauer, Martin Gottwald, and Klaus Diepold.
\newblock The successful ingredients of policy gradient algorithms.
\newblock In {\em Proceedings of the Thirtieth International Joint Conference
  on Artificial Intelligence, {IJCAI-21}}, pages 2455--2461. International
  Joint Conferences on Artificial Intelligence Organization, 8 2021.
\newblock Main Track.

\bibitem{henderson2018matters}
Peter Henderson, Riashat Islam, Philip Bachman, Joelle Pineau, Doina Precup,
  and David Meger.
\newblock Deep reinforcement learning that matters.
\newblock In {\em Proceedings of the Thirty-Second {AAAI} Conference on
  Artificial Intelligence}, pages 3207--3214. {AAAI} Press, 2018.

\bibitem{Hwangbo2019LearningAA}
Jemin Hwangbo, Joonho Lee, Alexey Dosovitskiy, Dario Bellicoso, Vassilios
  Tsounis, Vladlen Koltun, and Marco Hutter.
\newblock Learning agile and dynamic motor skills for legged robots.
\newblock {\em Science Robotics}, 4, 2019.

\bibitem{kaufmann2020RSS}
Elia Kaufmann, Antonio Loquercio, Ren{\'e} Ranftl, Matthias M{\"u}ller, Vladlen
  Koltun, and Davide Scaramuzza.
\newblock Deep drone acrobatics.
\newblock In {\em Proceedings of Robotics: Science and Systems}, Corvalis,
  Oregon, USA, July 2020.

\bibitem{kirk2021survey}
Robert Kirk, Amy Zhang, Edward Grefenstette, and Tim Rockt{\"{a}}schel.
\newblock A survey of generalisation in deep reinforcement learning.
\newblock {\em CoRR}, abs/2111.09794, 2021.

\bibitem{Kooi2021InclinedLanding}
Jacob~E. Kooi and Robert Babuska.
\newblock Inclined quadrotor landing using deep reinforcement learning.
\newblock In {\em {IEEE/RSJ} International Conference on Intelligent Robots and
  Systems, {IROS} 2021, Prague, Czech Republic, September 27 - Oct. 1, 2021},
  pages 2361--2368. {IEEE}, 2021.

\bibitem{Lambert2019LowLevelControl}
Nathan~O. Lambert, Daniel~S. Drew, Joseph Yaconelli, Sergey Levine, Roberto
  Calandra, and Kristofer S.~J. Pister.
\newblock Low-level control of a quadrotor with deep model-based reinforcement
  learning.
\newblock {\em IEEE Robotics and Automation Letters}, 4(4):4224--4230, 2019.

\bibitem{Lillicrap2016ContinuousControl}
Timothy~P. Lillicrap, Jonathan~J. Hunt, Alexander Pritzel, Nicolas Heess, Tom
  Erez, Yuval Tassa, David Silver, and Daan Wierstra.
\newblock Continuous control with deep reinforcement learning.
\newblock In Yoshua Bengio and Yann LeCun, editors, {\em 4th International
  Conference on Learning Representations, {ICLR} 2016, San Juan, Puerto Rico,
  May 2-4, 2016, Conference Track Proceedings}, 2016.

\bibitem{loquercio2019droneracing}
Antonio Loquercio, Elia Kaufmann, Ren{\'e} Ranftl, Alexey Dosovitskiy, Vladlen
  Koltun, and Davide Scaramuzza.
\newblock Deep drone racing: From simulation to reality with domain
  randomization.
\newblock {\em IEEE Transactions on Robotics}, 2019.

\bibitem{Lupashin2009SimpleLearningStrategy}
Sergei Lupashin, Angela Sch{\"o}llig, Michael Sherback, and Raffaello D'Andrea.
\newblock A simple learning strategy for high-speed quadrocopter multi-flips.
\newblock In {\em Proceedings of the IEEE International Conference on Robotics
  and Automation (ICRA) 2010}, Zurich, 2009.
\newblock 2010 IEEE International Conference on Robotics and Automation;
  Conference Location: Anchorage, AK, USA; Conference Date: May 3-8, 2010.

\bibitem{Mahony2012QuadrotorTutorial}
Robert Mahony, Vijay Kumar, and Peter Corke.
\newblock Multirotor aerial vehicles: Modeling, estimation, and control of
  quadrotor.
\newblock {\em IEEE Robotics Automation Magazine}, 19(3):20--32, 2012.

\bibitem{Mellinger2011MinimumSnap}
Daniel Mellinger and Vijay Kumar.
\newblock Minimum snap trajectory generation and control for quadrotors.
\newblock In {\em 2011 IEEE International Conference on Robotics and
  Automation}, pages 2520--2525, 2011.

\bibitem{mnih2015human}
Volodymyr Mnih, Koray Kavukcuoglu, David Silver, Andrei~A. Rusu, Joel Veness,
  Marc~G. Bellemare, Alex Graves, Martin Riedmiller, Andreas~K. Fidjeland,
  Georg Ostrovski, Stig Petersen, Charles Beattie, Amir Sadik, Ioannis
  Antonoglou, Helen King, Dharshan Kumaran, Daan Wierstra, Shane Legg, and
  Demis Hassabis.
\newblock Human-level control through deep reinforcement learning.
\newblock {\em Nature}, 518:529 EP --, 02 2015.

\bibitem{Molchanov2019Sim-to-Multi-Real}
Artem Molchanov, Tao Chen, Wolfgang H{\"{o}}nig, James~A. Preiss, Nora Ayanian,
  and Gaurav~S. Sukhatme.
\newblock Sim-to-(multi)-real: Transfer of low-level robust control policies to
  multiple quadrotors.
\newblock {\em CoRR}, abs/1903.04628, 2019.

\bibitem{panerati2021GymPyBulletDrones}
Jacopo Panerati, Hehui Zheng, SiQi Zhou, James Xu, Amanda Prorok, and Angela~P.
  Schoellig.
\newblock Learning to fly---a gym environment with pybullet physics for
  reinforcement learning of multi-agent quadcopter control.
\newblock In {\em 2021 IEEE/RSJ International Conference on Intelligent Robots
  and Systems (IROS)}, 2021.

\bibitem{Peng2018Sim2Real}
Xue~Bin Peng, Marcin Andrychowicz, Wojciech Zaremba, and Pieter Abbeel.
\newblock Sim-to-real transfer of robotic control with dynamics randomization.
\newblock In {\em 2018 IEEE International Conference on Robotics and Automation
  (ICRA)}, pages 3803--3810, 2018.

\bibitem{Schulman2016GAE}
John Schulman, Philipp Moritz, Sergey Levine, Michael Jordan, and Pieter
  Abbeel.
\newblock High-dimensional continuous control using generalized advantage
  estimation.
\newblock In {\em Proceedings of the International Conference on Learning
  Representations}, 2016.

\bibitem{schulman2017ppo}
John Schulman, Filip Wolski, Prafulla Dhariwal, Alec Radford, and Oleg Klimov.
\newblock Proximal policy optimization algorithms.
\newblock {\em CoRR}, abs/1707.06347, 2017.

\bibitem{Seyde2021bangBang}
Tim Seyde, Igor Gilitschenski, Wilko Schwarting, Bartolomeo Stellato, Martin
  Riedmiller, Markus Wulfmeier, and Daniela Rus.
\newblock Is bang-bang control all you need? solving continuous control with
  bernoulli policies.
\newblock In {\em Thirty-Fifth Conference on Neural Information Processing
  Systems}, 2021.

\bibitem{Silver2018AlphaZero}
David Silver, Thomas Hubert, Julian Schrittwieser, Ioannis Antonoglou, Matthew
  Lai, Arthur Guez, Marc Lanctot, Laurent Sifre, Dharshan Kumaran, Thore
  Graepel, Timothy Lillicrap, Karen Simonyan, and Demis Hassabis.
\newblock A general reinforcement learning algorithm that masters chess, shogi,
  and go through self-play.
\newblock {\em Science}, 362(6419):1140--1144, 2021/12/10 2018.

\bibitem{Tobin2017DomainRandomization}
Josh Tobin, Rachel Fong, Alex Ray, Jonas Schneider, Wojciech Zaremba, and
  Pieter Abbeel.
\newblock Domain randomization for transferring deep neural networks from
  simulation to the real world.
\newblock In {\em 2017 IEEE/RSJ International Conference on Intelligent Robots
  and Systems (IROS)}, pages 23--30, 2017.

\end{thebibliography}

\end{document}